\documentclass[letterpaper]{article} 
\usepackage{aaai24}  
\pdfoutput=1
\usepackage{times}  
\usepackage{helvet}  
\usepackage{courier}  
\usepackage[hyphens]{url}  
\usepackage{graphicx} 
\usepackage{amsmath}
\usepackage{natbib}  
\usepackage{caption} 
\usepackage{algorithm}
\usepackage{algorithmic}
\usepackage{booktabs}
\usepackage{multirow}
\usepackage{amssymb}
\usepackage{xcolor, colortbl}
\definecolor{Gray0}{gray}{0.95}
\definecolor{Gray}{gray}{0.85}
\definecolor{Gray1}{gray}{0.9}
\usepackage{bbding}
\usepackage{color}
\usepackage{graphicx}
\usepackage{float}
\usepackage{subfig}
\usepackage{newfloat}
\usepackage{listings}
\DeclareCaptionStyle{ruled}{labelfont=normalfont,labelsep=colon,strut=off} 
\floatstyle{ruled}
\newfloat{listing}{tb}{lst}{}
\floatname{listing}{Listing}
\title{Improved Face Representation via Joint Label Classification and Supervised Contrastive Clustering}
\author {
    Zhenduo Zhang
}
\affiliations{
    Platform Technology Department,  OVBU,  PCG, Tencent,  China  \\
    ericzdzhang@163.com
}

\usepackage{bibentry}

\begin{document}

\maketitle

\begin{abstract}
Face clustering tasks can learn hierarchical semantic information from large-scale data, which has the potential to help facilitate face recognition. However, there are few works on this problem. This paper explores it by proposing a joint optimization task of label classification and supervised contrastive clustering to introduce the cluster knowledge to the traditional face recognition task in two ways. We first extend ArcFace with a cluster-guided angular margin to adjust the within-class feature distribution according to the hard level of face clustering. Secondly,  we propose a supervised contrastive clustering approach to pull the features to the cluster center and propose the cluster-aligning procedure to align the cluster center and the learnable class center in the classifier for joint training. Finally, extensive qualitative and quantitative experiments on popular facial benchmarks demonstrate the effectiveness of our paradigm and its superiority over the existing approaches to face recognition.

\end{abstract}

\section{Introduction}
Face recognition is one of the most challenging tasks in pattern recognition and machine vision, and there are increasing demands in many industry areas. Current SOTA face recognition losses are mainly margin-based softmax losses~\cite{SphereFace, CosFace, Ad_loss, ArcFace, CircleLoss, CurricularFace, Sub-centerArcface, GroupFace, VPL-ArcFace}, which enforce greater intra-class compactness and inter-class discrepancy by adding an extra margin. They help pull face features towards the learnable class center with the same label to guide the label predicting. Except for the great development in face recognition, face clustering has made remarkable progress and can learn rich and hierarchical semantic information from large-scale data. Intuitively,  face clustering task~\cite{Kmeans, DBSCAN, L-GCN, V-GCN} may promote the classification task in traditional face recognition. To the best of our knowledge, how to incorporate information from clustering into the learning process of label prediction has been little explored in academia. 

Several works try to unify face recognition and face clustering. For instance, FaceNet~\cite{FaceNet} solves this by adopting the triplet metric loss to pull the sample towards the positive anchor and push it away from the negative anchor. However, FaceNet does not optimize the classification task directly,  while the current state-of-the-art methods' performances are achieved by classification. CenterLoss~\cite{CenterLoss} jointly optimizes the classification task by minimizing the Softmax loss and optimizing the clustering task by reducing the distance between the feature and its class center. However, CenterLoss conducts the joint learning in a naive way and only considers the intra-class compactness and ignores the inter-class discrepancy. MagFace~\cite{MegFace} points out that the magnitude of the face feature can be regarded as the indicator of face quality and can be used for face quality clustering. Furthermore,  face quality can guide the label prediction by adaptively adjusting the margin in the label classification loss.  However,  face quality is not the only influence factor on face recognition in real life scenarios and the hard samples are mainly due to large variations in pose,  age,  and occlusion. And these faces with large variations are usually harder to cluster, hence we should also pay attention to these samples with large variations instead of merely considering face quality. 

Recently,  contrastive learning has been employed in the clustering task and performs well on various benchmarks~\cite{PCL, CC}. In this work,  we try to explore the problem through the joint optimization task of face label prediction and supervised contrastive face clustering. We try to integrate the clustering information into the classic face recognition from two aspects. The level of clustering concentration depicts the compactness of the cluster result~\cite{PCL}. From the geometric perspective,  faces tend to be more easily pulled to the cluster center and more easily classified if the face cluster has a larger concentration. Thus,  in the first aspect,  we extend the ArcFace loss function with a cluster-guided angular margin to adaptively tune the classification decision boundary according to the hard level of clustering, and the loss is named as \textit{Cluster-Guided ArcFace(CG-ArcFace)}. In a nutshell,  the class with a larger concentration should be assigned with a smaller angular margin, and the class with a smaller concentration should be assigned with a larger angular margin. In this way, we can utilize the clustering information to facilitate the classification in an explicit manner compared with FaceNet and CenterLoss.  In the other aspect, to further use face clustering to promote face recognition,  we propose a supervised contrastive face clustering approach and a cluster-aligning procedure to jointly optimize both clustering and recognition tasks, which jointly considers the intra-class compactness and inter-class discrepancy.  The optimization of clustering helps the feature extractor to learn the prototype of each class at the feature level explicitly,  which helps to improve the robustness of the feature extractor to hard samples with large variations in pose, age, occlusion, etc.  Compared with the MagFace, we add an explicit clustering learning process to pull faces with different variations towards the cluster center.

In summary,  the contributions of this work are:

\begin{itemize}

\item We extend ArcFace with a cluster-guided angular margin to adjust the within-class feature distribution based on the cluster concentration,  which injects the cluster knowledge to label classification for joint learning.   

\item We further propose to jointly optimize both clustering and recognition tasks via a supervised contrastive face clustering approach and an additional cluster-aligning procedure. 

\item Extensive qualitative and quantitative results on popular facial benchmarks prove the effectiveness of our approach and the superiority over the existing methods to face recognition.
\end{itemize}

\section{Related Work}

\subsection{Face Recognition}
The current face recognition approaches can be divided into metric-learning methods and classification methods. In the metric-learning approaches,  FaceNet~\cite{FaceNet} minimizes the distance between the anchor and positive samples and maximizes the distance between the anchor and negative samples. CenterLoss~\cite{CenterLoss} minimizes the Euclidean distance between the face feature and its class center. In the classification category,  the typical loss functions include SphereFace~\cite{SphereFace}, CosFace~\cite{CosFace}, ArcFace~\cite{ArcFace} and so on. They enforce better intra-class compactness and inter-class discrepancy by adding an extra margin. They help pull face instances towards the learnable class center with the same identity, resulting in a discriminative face representation. Since a fixed margin in the margin-based loss leads the network to treat each sample equally without considering their importance degree, mining-based strategies are adopted to pay more attention to hard samples. MV-Softmax~\cite{MV-Softmax} and CurricularFace~\cite{CurricularFace} define hard samples as misclassified samples and integrate margin and mining into one framework. They emphasize hard samples by adopting a preset constant (MV-Softmax) or an adaptive variable (CurricularFace) as the weights of negative cosine similarities.
Besides,  AdaptiveFace~\cite{AdaptiveLoss},  AdaCos~\cite{AdaCos} and FairLoss~\cite{FairLoss} utilize adaptive margin strategy to automatically tune hyperparameters during training. MagFace~\cite{MegFace} is a joint framework of face recognition and clustering based on face quality. The feature magnitude is used as the quality indicator, and the adaptive margin and regularization item help enforce the face with higher quality to have a larger magnitude.

\subsection{Face Clustering}
The works in face clustering can be divided into two categories. Although promising results have been achieved,  classic unsupervised clustering gives discouraging results on large-scale complex datasets due to the naive distribution assumptions~\cite{Kmeans, DBSCAN}. Hence,  some supervised methods based on the graph convolutional network (GCN) have been proposed recently. For instance,  L-GCN~\cite{L-GCN} predicts the linkage on subgraphs deploying a GCN. DS-GCN~\cite{DS-GCN} and VE-GCN~\cite{VE-GCN} utilize two-stage GCNs to cluster the faces. STAR-FC~\cite{STAR-FC} proposes a structure-preserved sampling strategy to train the edge classification GCN.  

In our method, we try to cluster the faces in a supervised contrastive learning way, motivated by the works~\cite{SSC, PCL}. Instead of developing clustering methods,  our approach aims at improving feature distribution structure and providing the input feature for the mainstream clustering methods, which is similar to the MagFace~\cite{MegFace}.

\section{Proposed Approach}

\begin{figure*}
\centering
\includegraphics[width=0.6\linewidth]{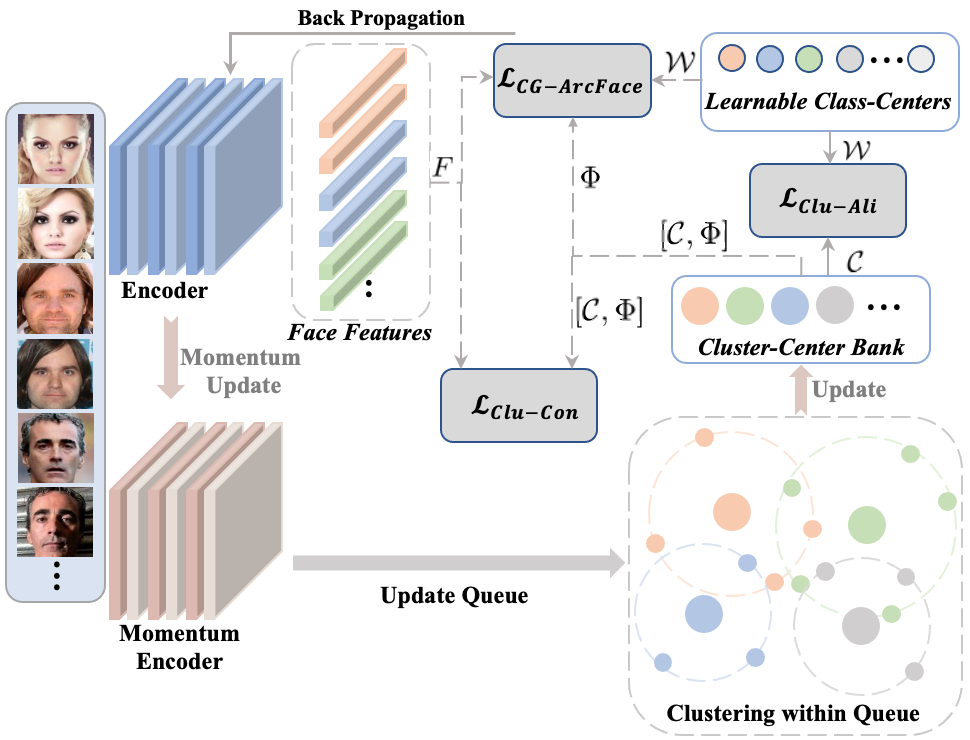}
\caption{An overview of the proposed approach. The framework contains an encoder and a momentum encoder. The Cluster-Guided ArcFace loss,  the supervised Cluster Contrastive loss, and the Cluster Aligning loss are utilized for training.}\label{fig:1}
\vspace{-1em}
\end{figure*}

Figure~\ref{fig:1} demonstrates the overall framework of our approach. The framework contains an encoder which is optimized by backpropagation and parameterized by $\theta_{f}$, and an momentum encoder parameterized by $\theta_{m}$ and updated by momentum:  $\theta_{m} \leftarrow m_{e}\theta_{m} + (1-m_{e})\theta_{f}$~\cite{MoCo}. $m_e$ is the momentum to control the updating speed of parameters. The momentum encoder generates face features to dynamically maintain a Feature Queue $Q$. We cluster face features according to the labels within the $Q$.  We denote the class number as $K$ and the class set of training dataset is denoted as $\mathcal{K} = \lbrace 1,2,...,K \rbrace$. The feature set of class $k$ in $Q$ is denoted as $Q_k$, and the cluster center of class $k$ within $Q$ is $C^{q}_k$.   $C^q_k$ is calculated by $C^q_k = \frac{1}{\vert Q_k \vert} \sum_{f \in Q_k} f$.  $C^q_k$ is used for updating the cluster center of class $k$ in the Cluster-Center Bank, which is denoted as $C_k$. The updating process is $C_k \leftarrow m_c C_k + (1-m_c)C^q_k$, where $m_c$ is the momentum controlling the updating speed.

Given the current $Q$ and the updated Cluster-Center Bank, we can calculate and update the concentration level for each class.  Referring the work~\cite{PCL}, the concentration is measured by the Equation~\ref{E1}, where \textbf{a smaller $\phi$ indicates larger concentration}. $\alpha$ is a smooth parameter to ensure that small clusters do not have an overly-large $\phi$.

\begin{small}
\begin{equation}
\begin{aligned}
\phi_k= \frac{\sum_{f \in Q_k} \Vert f-C_k \Vert_2}{|Q_k| log \left( |Q_k| + \alpha  \right) }, k \in \mathcal{K} = \lbrace 1,2,..., K \rbrace  \label{E1}
\end{aligned} 
\end{equation}
\end{small}

In the above way,  we can obtain each class's cluster center and corresponding cluster concentration.  Then we sample $M$ classes among all the $K$ classes, and the sampled cluster centers and cluster concentrations are denoted as $\Phi$ and $\mathcal{C}$ respectively.  The $\Phi$ and $\mathcal{C}$ are involved in loss function design afterward.

Three loss functions are employed in our framework, which are the Cluster-Guided ArcFace loss $\mathcal{L}_{CG-ArcFace}$,  the supervised Cluster Contrastive loss $\mathcal{L}_{Clu-Con}$ and the Cluster Aligning loss $\mathcal{L}_{Clu-Ali}$.

\subsection{Cluster-Guided ArcFace}

Before introducing our proposed Cluster-Guided ArcFace loss, we briefly revisit the ArcFace~\cite{ArcFace}. We suppose that we are given a training batch of $\mathcal{B}$ face samples $\lbrace f_i, y_i \rbrace^{\mathcal{B}}_{i=1}$, where $f_i \in R^d$ denotes the d-dimensional embedding and $y_i$ is its associated class label.  By defining the angle $\theta_{j}$ between $f_i$
and $j$-th  learnable class center $\mathcal{W}_j \in R^d$ as $\mathcal{W}^T_jf_i = \Vert \mathcal{W}_j \Vert \Vert f_i \Vert cos\theta_j$, the objective of ArcFace can be formulated as Equation~\ref{E2}:
\begin{small}
\begin{equation}
\begin{aligned}
\mathcal{L}_{ArcFace} = -\dfrac{1}{\mathcal{B}} \cdot \sum^{\mathcal{B}}_{i=1}  log \left(  \dfrac{e^{s\cdot cos(\theta_{y_{i}}+m)}}{e^{s\cdot cos(\theta_{y_{i}}+m)}+\sum_{j\neq y_{j}}e^{s\cdot cos \theta_j}} \right)  \label{E2}
\end{aligned} 
\end{equation}
\end{small}

where $m>0$ denotes the additive angular margin and $s$ is the scaling parameter. 

Despite its superior performance in enforcing intra-class compactness and inter-class discrepancy,  ArcFace employs the uniform margin $m$ for each class without considering the feature distribution structure.  Different from MagFace~\cite{MegFace} that utilizes the face quality to adjust the margin,  we directly explore the clustering structure knowledge and adjust the uniform margin based on the cluster concentration.  We suppose the class with a smaller concentration, i.e., larger $\phi$, should be assigned with a larger margin.  From the perspective of clustering, the smaller concentration indicates that faces are harder to cluster, and there are more hard samples in this class.  Assigning the class of smaller concentration with a larger margin will help pulling the hard samples towards the learnable class center and improves face recognition in the wild.  Hence, we reformulate the Equation~\ref{E2} to derive the Cluster-Guided ArcFace loss.  The $\mathcal{L}_{CG-ArcFace}$ is formulated in Equation~\ref{E3} and Equation~\ref{E4}.

\begin{small}
\begin{equation}
\begin{aligned}
\mathcal{L}_{CG-ArcFace} = -\dfrac{1}{\mathcal{B}} \cdot \sum^{\mathcal{B}}_{i=1}  \mathcal{L}_i  \label{E3}
\end{aligned} 
\end{equation}
\end{small}

\begin{small}
\begin{equation}
\begin{aligned}
\mathcal{L}_i = log \left(  \dfrac{e^{s\cdot cos(\theta_{y_{i}}+\lambda(\phi_{y_i})\cdot m)}}{e^{s\cdot cos(\theta_{y_{i}}+\lambda(\phi_{y_i})\cdot m)}+\sum_{j\neq y_{j}}e^{s\cdot cos \theta_j}} \right)  \label{E4}
\end{aligned} 
\end{equation}
\end{small}

where $\lambda(\phi_{y_i})$ is a scale factor to multiply the uniform margin $m$ and has the property of monotonically increasing with the $\phi_{y_i}$ and monotonically decreasing with the concentration.  Hence,  if the class $y_i$ has smaller concentration, a larger margin $\lambda(\phi_{y_i})\cdot m$ should be assigned to it.

We can define $\lambda(\phi_{y_i})$ in a linear manner. We denote the maximum concentration of all classes as $\phi_{max} = max \lbrace \phi_{k}, k \in \mathcal{K} \rbrace$ and the minimum concentration as $\phi_{min} = min \lbrace \phi_{k}, k \in \mathcal{K} \rbrace$.  Then $\lambda(\phi_{y_i})$ is defined as Equation~\ref{E5}.
\begin{small}
\begin{equation}
\begin{aligned}
\lambda(\phi_{y_i}) = \frac{\phi_{y_i} - \phi_{min}}{\phi_{max} - \phi_{min}} \label{E5}
\end{aligned} 
\end{equation}
\end{small}

\subsection{Supervised Contrastive Clustering}
Except for adjusting the classification margin based on the cluster concentration, which is a manner of joint learning of classification and clustering,  we furthermore propose to jointly optimize both tasks via a supervised contrastive clustering process and an additional clustering-aligning process.  

We employ the infoNCE~\cite{infoNCE} loss to conduct contrastive learning of the instance features and cluster centers,  named as supervised Cluster Contrastive loss, $\mathcal{L}_{Clu-Con}$.  Given an instance feature,  we can obtain the positive cluster center according to its label and obtain some other cluster centers as negative samples.  We regard it as a supervised contrastive manner because we extend the traditional self-supervised batch contrastive approach to the fully-supervised setting with reference to SupCon~\cite{SSC}.

Assuming that the input mini-batch contains $\mathcal{B}$ samples and the number of sampled cluster centers is $M$,  we denote the features and class labels of the mini-batch as $F = \left[ f_1, f_2, ..., f_{\mathcal{B}} \right]$  and $\left[ y_1, y_2, ..., y_{\mathcal{B}} \right],  y_i \in \mathcal{K}$.  The sampled cluster centers are denoted as $\mathcal{C}=\left[ C^s_1, C^s_2, ..., C^s_M \right]$, and their labels and concentration measurements are $\left[ y^s_1, y^s_2, ..., y^s_M \right]$ and $\Phi=\left[ \phi^s_1, \phi^s_2, ..., \phi^s_M \right]$. Then the supervised Cluster Contrastive loss is defined as Equation~\ref{E6}, where $1\left( y_i = y^s_j\right)  = 1$ if $y_i$ is the same as $y^s_j$; otherwise $1\left( y_i = y^s_j\right)  = 0$.

\begin{small}
\begin{equation}
\begin{aligned}
\mathcal{L}_{Clu-Con} = -\frac{1}{\mathcal{B}}\sum^{\mathcal{B}}_{i=1} \log \frac{\sum^{M}_{j=1}\exp\left(  f_i \cdot C^s_j / \phi^s_j \right) 1\left( y_i = y^s_j\right)  }{\sum^{M}_{j=1}\exp\left(  f_i \cdot C^s_j / \phi^s_j \right) }  \label{E6}
\end{aligned} 
\end{equation}
\end{small}

The class-adaptive temperature $\phi^s_j$ is designed with reference to PCL~\cite{PCL}.  It helps make $\mathcal{L}_{Clu-Con}$ be adaptive to samples with different hardness levels.

\subsubsection{Major Differences from the Related Work} The supervised Cluster Contrastive loss is designed with reference to SupCon~\cite{SSC} and PCL~\cite{PCL}. SupCon first proposes to conduct contrastive learning in a supervised way, and the positive sample is selected from the images with the same class label. Our work inherits the idea of supervised learning from SupCon, and the main difference is that we choose the cluster centers rather than images to form positive and negative samples and adopt a class-adaptive temperature for face clustering. PCL proposes to utilize class prototypes to bridge contrastive learning and clustering while the class prototypes in their work are generated using the Expectation-Maximization(EM)-based algorithm,  which is also an unsupervised learning way. The main difference between our work and PCL is that our cluster centers are generated in a supervised manner and make full usage of the face labels. 

In addition to the contrastive learning between cluster centers and instances,  we add an extra contrastive learning process between cluster centers and the learnable class centers in the label classifier, and we employ the cluster-aligning loss,  $\mathcal{L}_{Clu-Ali}$ to implement it. $\mathcal{L}_{Clu-Ali}$ will help align the cluster center $\mathcal{C}_{y_i}$ and the corresponding learnable class center $\mathcal{W}_{y_i}$, and this may promote the classifier optimization using the clustering results. We define the $\mathcal{L}_{Clu-Ali}$ as Equation~\ref{E7}, where $\tau$ is the learnable temperature widely used in contrastive learning.

\begin{small}
\begin{equation}
\begin{aligned}
\mathcal{L}_{Clu-Ali} = -\frac{1}{M}\sum^{M}_{i=1} \log \frac{\exp\left(C^s_i \cdot  \mathcal{W}_{y^s_i}  / \tau \right)  }{\sum^{K}_{j=1}\exp\left( C^s_i \cdot \mathcal{W}_j  / \tau \right) }  \label{E7}
\end{aligned} 
\end{equation}
\end{small}

\subsection{Overall Loss}
The overall loss of our framework is written as Equation.~\ref{E8}, where $\lambda_1$ and $\lambda_2$ are the weights.

\begin{equation}
\begin{aligned}
\mathcal{L} = \mathcal{L}_{CG-ArcFace} +  \lambda_1 \mathcal{L}_{Clu-Con} + \lambda_2 \mathcal{L}_{Clu-Ali}  \label{E8}
\end{aligned} 
\end{equation}

\section{Experiments}
\subsection{Implementation Details}
We utilize the refined MS1M~\cite{MS1M} as our training dataset to conduct a fair comparison with other methods. In the testing stage, we extensively evaluate our
approach on popular benchmarks, including LFW~\cite{LFW}, CFP-FP~\cite{CFP-FP}, CPLFW~\cite{CPLFW}, AgeDB~\cite{AgeDB}, CALFW~\cite{CALFW}, IJB-B~\cite{IJB-B}, IJB-C~\cite{IJB-C} and MegaFace~\cite{MegaFace}. For data pre-processing, we first resize the aligned face images to $112 \times 112$. For the selection of backbone networks, we use the most widely used CNN architectures ResNet~\cite{ResNet}. All experiments in this paper are implemented using PyTorch, and we will release our code and pretrained models in the near future.

The batch size $\mathcal{B}$ is set to 512, and models are trained on 8 NVIDIA Tesla V100 GPUs. We employ the Adam optimizer in the training stage, and the learning rate starts from 0.001. We decrease the learning rate by $0.1 \times $ at $20_{th}$,  $40_{th}$, and $60_{th}$ epochs and stop at $80_{th}$ epochs. The momentum $m_e$ for updating the momentum encoder is set to 0.999, and the size of the Feature Queue $Q$ is set to 8192.  The momentum $m_c$ for updating the Cluster-Center Bank is set to 0.9. The number of the class centers, $M$ in Equation~\ref{E6} and Equation~\ref{E7}, is 2048. The margin parameter $m$ and the scale parameter $s$ in Equation~\ref{E4} are 0.5 and 64.  The smooth parameter in Equation~\ref{E1} is set to 10. The loss weights $\lambda_1$ and $\lambda_2$ are 1.0 and 0.5, respectively. 

\subsection{Comparison with the State-of-the-Art Methods}

\begin{table*}[ht]
\normalsize
\setlength{\abovecaptionskip}{0cm}
\setlength{\belowcaptionskip}{0cm}
\centering

 \resizebox{\linewidth}{!}{
\begin{tabular}{c|c|c|c|c|c|c|c|c|c }
\toprule[1pt]
\multirow{2}{*}{Methods} &\multicolumn{5}{c|}{Verification Accuracy} &\multicolumn{2}{c|}{IJB} & \multicolumn{2}{c}{MegaFace} \\
\cline{2-10}
& LFW & CFP-FP& CPLFW& AgeDB& CALFW & IJB-B & IJB-C & Id & Ver \\
\hline
CosFace\cite{CosFace}(CVPR18) & $99.81$ &$98.12$ & $92.28$& $98.11$& $95.76$ & $94.80$ & $96.37$ & $97.91$ & $97.91$ \\
ArcFace\cite{ArcFace}(CVPR19) & $99.83$ &$98.27$ & $92.08$& $98.28$& $95.45$ & $94.25$ & $96.03$ & $98.35$ & $98.48$ \\
AFRN\cite{AFRN} (ICCV19)& $99.85$ &$95.56$ & $93.48$& $95.35$& $\mathbf{96.30}$ & $88.50$ & $93.00$ & $-$ & $-$ \\
MV-Softmax\cite{MV-Softmax}(AAAI20) & $99.80$ &$98.28$ & $92.83$& $97.95$& $96.10$ & $93.60$ & $95.20$ & $97.76$ & $97.80$ \\
GroupFace\cite{GroupFace} (CVPR20)& $99.85$ &$98.63$ & $93.17$& $98.28$& $96.20$ & $94.93$ & $93.26$ & $98.74$ & $98.79$ \\
CircleLoss\cite{CircleLoss}(CVPR20) & $99.73$ &$96.02$ & $-$& $-$& $-$ & $-$ & $93.95$ & $98.50$ & $98.73$ \\
DUL\cite{DUL}(CVPR20) & $99.83$ &$98.78$ & $-$& $-$& $-$ & $-$ & $94.61$ & $98.60$ & $-$ \\
CurricularFace\cite{CurricularFace} (CVPR20)& $99.80$ &$98.37$ & $93.13$& $98.32$& $96.20$ & $94.80$ & $96.10$ & $98.71$ & $98.64$ \\
URFace\cite{URFace}(CVPR20) & $99.78$ &$98.64$ & $-$& $-$& $-$ & $-$ & $96.60$ & $-$ & $-$ \\
DB\cite{DB}(CVPR20) & $99.78$ &$-$ & $92.63$& $97.90$& $96.08$ & $-$ & $-$ & $96.35$ & $96.56$ \\
Sub-center ArcFace\cite{Sub-centerArcface}(ECCV20) & $99.80$ &$98.80$ & $-$& $98.31$& $-$ & $94.94$ & $96.28$ & $98.16$ & $98.36$ \\
BroadFace\cite{BroadFace}ECCV20 & $99.85$ &$98.63$ & $93.17$& $98.38$& $96.20$ & $94.97$ & $96.38$ & $98.70$ & $98.95$ \\
SST\cite{SST}ECCV20 & $99.75$ &$95.10$ & $88.35$& $97.20$& $94.92$ & $-$ & $-$ & $96.27$ & $96.96$ \\
VPL-ArcFace\cite{VPL-ArcFace} (CVPR21) & $99.83$ &$99.11$ & $93.45$& $\mathbf{98.60}$& $96.12$ & $95.56$ & $96.76$ & $\mathbf{98.80}$ & $\mathbf{98.98}$ \\
ElasticFace\cite{ElasticFace} (CVPRW22) & $99.82$ &$98.60$ & $93.28$& $98.35$& $96.17$ & $95.09$ & $96.40$ & $\mathbf{98.80}$ & $98.83$ \\
\hline
\rowcolor{Gray}
Ours(ResNet-100) & $\mathbf{99.85}$ &$\mathbf{99.12}$ & $\mathbf{93.51}$& $98.58$& $96.16$ & $\mathbf{95.62}$ & $\mathbf{96.80}$ & $\mathbf{98.80}$ & $98.95$ \\
\bottomrule[1pt]
\end{tabular}
}
\caption{Performance comparisons with the state-of-the-art methods on various benchmarks. 1:1 verification
accuracy ($\%$) is reported on the LFW, CFP-FP, CPLFW, AgeDB, CALFW datasets. TAR@FAR=1e-4 is reported on the IJB-B and
IJB-C datasets. Identification and verification evaluation on MegaFace using FaceScrub as the probe set. “Id” refers to the rank-1 face
identification accuracy with 1M distractors, and “Ver” refers to the face verification TAR@FPR=1e-6.}
\label{table:SOTA}
\vspace{0em}
\end{table*}

To compare with recent state-of-the-art competitors, we train our model on the MS1M dataset, and the backbone we adopt is ResNet-100 for a fair comparison. Our model is tested on various benchmarks, including LFW for unconstrained face verification, CFP-FP and CPLFW for large pose variations, AgeDB and CALFW for age variations, IJB-B, and IJB-C for mixed-media (image and video) face verification, and MegaFace for identification and verification under million-scale distractors. As is reported in Table~\ref{table:SOTA},  the proposed method achieves the state-of-the-art result ($99.85$) with the competitors on LFW, where the performance is almost saturated. For pose-invariant and age-invariant face recognition, our method achieves $99.12\%$ on CFP-FP, $93.51\%$ on CPLFW,  $98.58\%$ on AgeDB, and $96.16\%$ on CALFW and outperforms most of the other state-of-the-art methods, including GroupFace, CurricularFace,  Sub-center ArcFace, BroadFace,  VPL-ArcFace, ElasticFace and etc. Besides, our model obtains $95.62$ TAR on IJB-B and $96.80$ TAR on IJB-C when FAR is set as $1e^{-4}$ and achieves the best performance on these two benchmarks. On MegaFace, our approach also has comparable results with the state-of-the-art method(VPL-ArcFace). Overall speaking,  the superiority of our algorithm has been well demonstrated by the comparison with the state-of-the-art approaches.

\subsection{Ablation Study}

\begin{table*}[ht]
\normalsize
\setlength{\abovecaptionskip}{0cm}
\setlength{\belowcaptionskip}{0cm}
\centering

 \resizebox{\linewidth}{!}{
\begin{tabular}{c|c|c|c|c|c|c|c|c|c|c|c|c }
\toprule[1pt]
\multirow{2}{*}{Settings} & \multicolumn{3}{c|}{Loss Functions} &\multicolumn{5}{c|}{Verification Accuracy} &\multicolumn{2}{c|}{IJB} & \multicolumn{2}{c}{MegaFace} \\
\cline{2-13}
& $\mathcal{L}_{CG-ArcFace}$& $\mathcal{L}_{Clu-Con}$ &$\mathcal{L}_{Clu-Ali}$ & LFW & CFP-FP& CPLFW& AgeDB& CALFW & IJB-B & IJB-C & Id & Ver \\
\hline

\textit{\textcolor{red}{baseline}} & &&&$99.83$ &$98.27$ & $92.08$& $98.28$& $95.45$ & $94.25$ & $96.03$ & $98.35$ & $98.48$ \\
\rowcolor{Gray0}
\textit{\textcolor{blue}{setting1}}& \checkmark &  &&$99.82$ &$98.78$ & $92.94$& $98.43$& $95.80$ & $95.12$ & $96.38$ & $98.48$ & $98.64$ \\
\rowcolor{Gray0}
\textit{\textcolor{blue}{setting2}} & &\checkmark &&$99.85$ &$98.96$ & $93.26$& $98.50$& $95.96$ & $95.48$ & $96.68$ & $98.60$ & $98.72$ \\
\rowcolor{Gray0}
\textit{\textcolor{blue}{setting3}} & &&\checkmark  &$99.85$ &$98.36$ & $92.24$& $98.32$& $95.51$ & $94.80$ & $96.28$ & $98.38$ & $98.50$ \\
\rowcolor{Gray}
\textit{\textcolor{blue}{setting4}} & \checkmark & \checkmark &  &$99.83$ &$99.06$ & $93.45$& $98.52$& $96.12$ & $95.56$ & $96.75$ & $98.72$ & $98.88$ \\
\rowcolor{Gray}
\textit{\textcolor{blue}{setting5}} & \checkmark &  & \checkmark &$99.83$ &$98.85$ & $93.06$& $98.48$& $95.92$ & $95.27$ & $96.52$ & $98.54$ & $98.68$ \\
\rowcolor{Gray}
\textit{\textcolor{blue}{setting6}} & \checkmark & \checkmark & \checkmark &$99.85$ &$99.12$ & $93.51$& $98.58$& $96.16$ & $95.62$ & $96.80$ & $98.80$ & $98.95$ \\

\bottomrule[1pt]
\end{tabular}
}
\caption{Ablation studies of the improvements of different loss functions.  The performance indicators are explained in the caption of Table~\ref{table:SOTA}. The model of the first line without $\mathcal{L}_{CG-ArcFace}$,  $\mathcal{L}_{Clu-Con}$ and $\mathcal{L}_{Clu-Ali}$, is the baseline model, which uses ResNet-100 as backbone and the ArcFace loss as the classification loss.}
\label{table:ablation}
\vspace{-1em}
\end{table*}

In order to demonstrate the effectiveness and performance of our proposed method, we carried out sufficient ablation experimental studies to explore the contributions of the Cluster-Guided ArcFace $\mathcal{L}_{CG-ArcFace}$,  the supervised Cluster Contrastive loss $\mathcal{L}_{Clu-Con}$ and the Cluster-Aligning loss $\mathcal{L}_{Clu-Ali}$. Since our proposed classification loss, the Cluster-Guided ArcFace,  is based on the original ArcFace loss, we select the model with ArcFace classification head and ResNet-100 backbone as the \textcolor{red}{\textit{baseline}} model. The ablation studies are divided into two parts, where one part explores the performance improvement of individual loss functions, and the other one explores the joint performance improvement of multiple losses. The overall ablation studies are demonstrated in Table~\ref{table:ablation}.

\subsubsection{Performance Improvement of Individual Losses}
Firstly,  we study the effect of each loss function on the baseline separately.  

\textcolor{blue}{\textit{setting1:}}  We replace the ArcFace loss with the proposed Cluster-Guided ArcFace $\mathcal{L}_{CG-ArcFace}$ to utilize the cluster concentration of each class to adjust the decision margin adaptively.  The \textit{setting1} model significantly outperforms the \textit{baseline} model on all the benchmarks except for the LFW benchmark, where the performance is almost saturated. Especially, the \textit{setting1} model has a significant improvement on the benchmarks with hard face samples, including CFP-FP and CPLFW with large pose variations, AgeDB and CALFW with age variations,  IJB-B(C) containing images and frames from videos, and the MegaFace containing massive samples and a high degree of variability in scale, pose and occlusion. The hard samples with large variations in pose, age, scale, and occlusion, are usually distributed far away from the cluster center and are harder to cluster. We assume that the margin for their class should be tuned larger to help pulling the hard samples towards the learnable class center. The comparisons between the \textit{setting1} and the \textit{baseline} have proved the correctness of our assumption and the validity of the Cluster-Guided ArcFace.

\textcolor{blue}{\textit{setting2:}} We add the supervised Cluster Contrastive loss $\mathcal{L}_{Clu-Con}$ to the \textit{baseline} model to verify the effect of the joint optimization of both clustering and recognition task. From Table~\ref{table:ablation}, the \textit{setting2} model notably exceeds the \textit{baseline} on all the benchmarks especially the benchmarks with hard samples.  We believe that the experimental results can be interpreted as the optimization of clustering being able to help the feature extractor learn the prototype of each class at the feature level explicitly.  However, in ArcFace, the concept of the prototype is reflected through the classifier weights. The joint optimization of clustering and recognition is, in fact, a fusion of the two ideas of the learning prototype. Joint learning of class prototypes helps to improve the robustness of feature extraction of hard samples with large variations in pose,  age,  scale,  occlusion, etc.

\textcolor{blue}{\textit{setting3:}} We add the Cluster-Aligning loss $\mathcal{L}_{Clu-Ali}$ to facilitate the learning of ArcFace in the \textit{setting3}.  There are some improvements on different benchmarks, but the improvements are limited compared with the \textit{setting1} and \textit{setting2}.  The Cluster-Aligning loss can enhance the performance because the alignment between learnable class centers in the classifier and cluster centers helps the learning of the classifier to be robust to the variations of faces.  The reason why the improvements are limited is that the Cluster-Aligning loss serves a similar role as the ArcFace, and little additional knowledge is learned.

\subsubsection{Performance Improvement of Joint Combination of Multiple Losses}
In the above section, we have evaluated the individual improvement of the different losses, and all of them can promote face recognition performance.  We here examine the performance of the joint combination of them.  The comparisons between the settings (\textcolor{blue}{\textit{setting4}}, \textcolor{blue}{\textit{setting5}}) and \textit{setting1} demonstrate that both the supervised Cluster Contrastive loss and the Cluster-Aligning loss can further promote the performance of the Cluster-Guided ArcFace. Furthermore, the comparisons between \textcolor{blue}{\textit{setting6}} and the settings(\textit{setting4}, \textit{setting5}) prove that the supervised Cluster Contrastive loss and the Cluster-Aligning loss can promote each other.This is because the cluster aligning procedure helps to constrain the consistency of the cluster centers and the learnable centers, which should be consistent in theory.  In summary,  the comparisons between experimental settings have illustrated the effectiveness of joint label classification and supervised contrastive clustering and the contributions of the proposed loss functions.

\subsection{Comparisons with the Triplet Loss and Center Loss }
The supervised Cluster Contrastive loss has shown consistency in improving the performance of face recognition via clustering the face features iteratively. In comparison, some metric losses, such as the Triplet loss~\cite{FaceNet} and the Center loss~\cite{CenterLoss} can also play a similar role. We have conducted experiments to compare our supervised Cluster Contrastive loss, Triplet loss, and Center loss.

\begin{table}[ht]
\normalsize
\setlength{\abovecaptionskip}{0cm}
\setlength{\belowcaptionskip}{0cm}
\centering
 \resizebox{\linewidth}{!}{
\begin{tabular}{c|c|c|c|c|c }
\toprule[1pt]
\multirow{2}{*}{Losses} & \multicolumn{3}{c|}{Verification Accuracy} &\multicolumn{2}{c}{IJB} \\
\cline{2-6}
&  LFW & CFP-FP&  AgeDB&  IJB-B & IJB-C  \\
\hline
Center Loss & $99.78$ &$98.68$ & $98.36$&  $95.04$ & $96.34$  \\
Triplet Loss & $99.82$ &$98.89$ & $98.48$&  $95.32$ & $96.56$  \\ 
\rowcolor{Gray}
$\mathcal{L}_{Clu-Con}$ & $99.85$ &$99.12$ & $98.58$&  $95.62$ & $96.80$  \\

\bottomrule[1pt]
\end{tabular}
}
\caption{Comparison between the supervised Cluster Contrastive loss and the Triplet loss and Center Loss.  The performance indicators are explained in the caption of Table~\ref{table:SOTA}. }
\label{table:c1}
\vspace{-0.5em}
\end{table}

For a fair comparison, we only replace the supervised Cluster Contrastive loss with the Center Loss and the Triplet loss and still utilize the Cluster-Guided ArcFace loss and the Cluster-Aligning loss. The results in Table~\ref{table:c1} show that the proposed supervised Cluster Contrastive loss performs better than the Center Loss and the Triplet Loss in the five benchmarks, especially in the ones with more difficult samples(CFP-FP, AgeDB, IJB-B and IJB-C). Compared with the Center loss, the supervised Cluster Contrastive loss joint optimizes both the intra-class compactness and the inter-class discrepancy. Compared with the Triplet loss, the supervised Cluster Contrastive loss can meet more negative samples and is adaptive to the hard level of clustering.

\subsection{Effect of the Adaptive Temperature in  $\mathcal{L}_{Clu-Con}$}

We verify the effect of the cluster-adaptive temperature $\phi^s_j$
in the supervised Cluster Contrastive loss is adjusted according to the feature distribution in the cluster. We compare it with the typical learnable temperature $\tau_0$, widely adopted in the infoNCE~\cite{infoNCE} loss, on various benchmarks. As is illustrated in Table~\ref{table:c2}, the models using the cluster-adaptive temperature consistently outperform those using the typical learnable temperature on all benchmarks. This proves it will promote better face representation learning via making the temperature adaptive to the hard level of face clustering.

\begin{table}[ht]
\normalsize
\setlength{\abovecaptionskip}{0cm}
\setlength{\belowcaptionskip}{0cm}
\centering

 \resizebox{\linewidth}{!}{
\begin{tabular}{c|c|c|c|c|c }
\toprule[1pt]
\multirow{2}{*}{Temperature} & \multicolumn{3}{c|}{Verification Accuracy} &\multicolumn{2}{c}{IJB} \\
\cline{2-6}
&  LFW & CFP-FP&  AgeDB&  IJB-B & IJB-C  \\
\hline
Typical  $\tau_{0}$ & $99.83$ &$99.06$ & $98.50$&  $95.54$ & $96.72$  \\
\rowcolor{Gray}
Adaptive $\phi^s_j$ & $99.85$ &$99.12$ & $98.58$&  $95.62$ & $96.80$  \\

\bottomrule[1pt]
\end{tabular}
}
\caption{Effect of the Adaptive Temperature in  $\mathcal{L}_{Clu-Con}$.  The performance indicators are explained in the caption of Table~\ref{table:SOTA}. }

\label{table:c2}
\vspace{-1.5em}
\end{table}

\subsection{Comparison of Feature Similarity Distribution}

\begin{figure*}[!htb]
  \centering
  \subfloat[CFP-FP - ArcFace]{\includegraphics[trim=0 0 0 0,clip, width=0.37\textwidth]{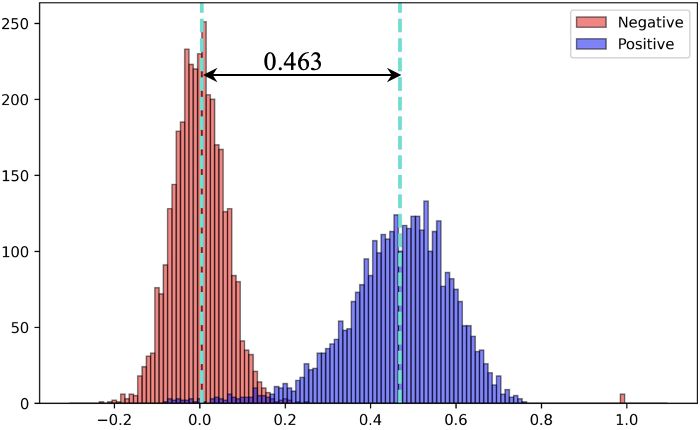}}   \subfloat[CFP-FP - Ours]{\includegraphics[trim=0 0 0 0,clip,width=0.37\textwidth]{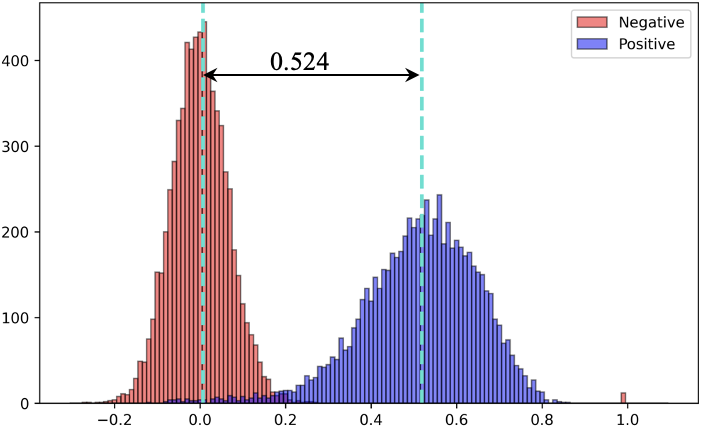}}\\
  \vspace{-5pt}
  \subfloat[AgeDB - ArcFace]{\includegraphics[trim=0 0 0 0,clip, width=0.37\textwidth]{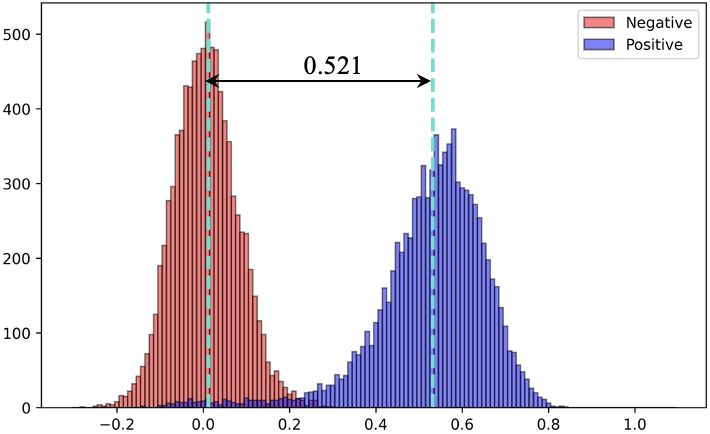}}  \subfloat[AgeDB - Ours]{\includegraphics[trim=0 0 0 0,clip, width=0.37\textwidth]{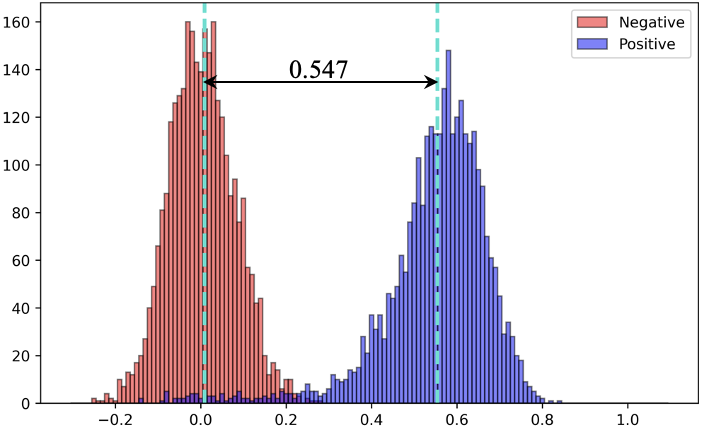}}\\
 
  \caption{Feature similarity distributions of ArcFace and our approach. The dashed line indicates the distribution expectation. }
  \label{fig:2}
  \vspace{-1em}
\end{figure*}

We also conduct a quantitive experiment to prove the effectiveness of our approach. We illustrate the feature similarity distribution on the CFP-FP and the AgeDB dataset. The red bars show the cosine similarity distribution of negative pairs, and the blue bars show the cosine similarity distribution of positive pairs. The vertical dashed line indicates the expectation of cosine similarity distribution. The margin between the two vertical dashed lines can indicate the ability of face verification. 

Figure 2(a) and Figure 2(b), the margin between positive pairs and negative pairs on the CFP-FP dataset is enlarged from $0.463$ to $0.524$, which shows that our paradigm can improve the face representation of the samples with large pose variations. Similarly,  the margin between positive pairs and negative pairs on the AgeDB dataset is enlarged from $0.521$ to $0.547$, shown in Figure 2(c) and Figure 2(d). This result proves that our paradigm can also improve the face representation of the samples with large age variations. The quantitive experiment further validates the effect of our paradigm on learning a better face representation.

\subsection{Face Clustering}

\setlength{\tabcolsep}{4pt}
\begin{table}
\small
\setlength{\abovecaptionskip}{0.0cm}
\setlength{\belowcaptionskip}{-0.0cm}

  \begin{center}
    \resizebox{0.48\textwidth}{!}{%
      \begin{tabular}{lccccccc}
       \toprule[1pt]
        Method & Net & \multicolumn{2}{c}{IJB-B-512} & \multicolumn{2}{c}{IJB-B-1024} & \multicolumn{2}{c}{IJB-B-1845}  \\
        \cline{3-8}
               & & F & NMI & F & NMI & F & NMI \\
        \hline
        \hline
        K-means & ArcFace & 66.70 & 88.83 & 66.82 & 89.48 &  66.93 & 89.88 \\
           \cellcolor{white} &  MagFace & 66.75 & 88.86 & 67.33& 89.62 &  67.06 & 89.96 \\
            \rowcolor{Gray1}   \cellcolor{white}        & Ours & \textbf{66.79}& \textbf{88.89} & \textbf{67.51}& \textbf{89.70} &  \textbf{67.12}& \textbf{90.02} \\
        
        DBSCAN & ArcFace & 72.72 & 90.42 & 72.50 & 91.15 &  73.89 & 91.96 \\
        \cellcolor{white}        & MagFace & 73.13& 90.61 & 72.68& 91.30&  74.26& 92.13 \\
        
        \rowcolor{Gray1}   \cellcolor{white}        & Ours & \textbf{73.44}& \textbf{90.75} & \textbf{72.78}& \textbf{91.42} &  \textbf{74.49}& \textbf{92.25} \\
        L-GCN & ArcFace & 84.92 & 93.72 & 83.50 & 93.78 & 80.35 & 92.30 \\
        \cellcolor{white}  & MagFace & 85.27 & 93.83 & 83.79 & 94.10 &  81.58 &  92.79 \\
        \rowcolor{Gray1}   \cellcolor{white}        & Ours & \textbf{85.46}& \textbf{93.88} & \textbf{83.92}& \textbf{94.41} &  \textbf{82.17}& \textbf{93.15} \\
       \bottomrule[1pt]
      \end{tabular}
    }
  \end{center}
  \caption{F-score (\%) and NMI (\%) on clustering benchmarks.} \label{table:clustering}
\vspace{-1em}
\end{table}
\setlength{\tabcolsep}{1.4pt}

Similar to MagFace~\cite{MegFace}, our approach aims to improve feature distribution structure and provide the input feature for the mainstream clustering methods. Hence, we compare the performance of our approach with the ones of the baseline model(ArcFace) and the recent work MagFace,  via integrating their features with multiple clustering methods. We utilize three clustering methods for evaluation: K-means~\cite{Kmeans}, DBSCAN~\cite{DBSCAN} and L-GCN~\cite{L-GCN}. Following the IJB-B clustering protocol, we evaluate on three largest sub-tasks where the numbers of identities are 512, 1024, and 1845. Normalized mutual information (NMI) and BCubed F-measure~\cite{10.1007/s10791-008-9066-8} are employed as the evaluation metrics. Following the IJB-B clustering protocol~\cite{IJB-B} we evaluate on three largest sub-tasks where the numbers of identities are 512, 1024, and 1845. Normalized mutual information (NMI) and BCubed F-measure are employed as the evaluation metrics.

Table~\ref{table:clustering} illustrates the clustering results. The overall performance can be consistently improved with stronger cluster methods(K-means$<$DBSCAN$<$L-GCN) compared with the baseline ArcFace model. Our approach improves both F-score and NMI metrics over the baseline model, which shows that our paradigm that jointly conducts label prediction and supervised contrastive clustering can promote the clustering performance.  We also compare our approach with the MagFace and our methods outperforms MagFace consistently.  Moreover, this may be contributed to that our Cluster-Guided ArcFace directly utilizes the cluster concentration, which contains more knowledge than face quality,  to guide the tuning of margin. On the other hand,  our supervised cluster contrastive loss conducts an explicit clustering procedure to help learning better face representation. The clustering results further prove that our approach can improve the feature representation.

\section{Conclusion}

In summary,   we propose a novel paradigm that jointly conducts label classification and face clustering to introduce cluster knowledge to the recognition task.  This paradigm is implemented from two aspects.  Firstly, we extend the ArcFace to Cluster-Guided ArcFace with a cluster-guided angular margin to adaptively tune the classification boundary according to the cluster concentration.  Secondly, we propose a supervised contrastive face clustering and cluster-aligning approach to jointly optimize the clustering and recognition tasks.  The comparisons with the state-of-the-art works show that our approach is superior to the existing methods.  The ablation studies sufficiently prove the validity of the proposed losses and learning paradigm.  The feature similarity distribution and face clustering results demonstrate that our method can improve the face representation from both the verification and clustering perspectives.  The quantitive and qualitative results are consistent with our basic assumptions.

\bibliography{aaai24}

\end{document}